\documentclass[nohyperref]{article}

\usepackage{microtype}
\usepackage{graphicx}
\usepackage{subfigure}
\usepackage{multirow}
\usepackage{placeins}
\usepackage{float}
\usepackage{booktabs} 
\usepackage[normalem]{ulem}

\usepackage{hyperref}

\usepackage[accepted]{styles/icml2023}

\usepackage{amsmath}
\usepackage{amssymb}
\usepackage{mathtools}
\usepackage{amsthm}

\usepackage[capitalize,noabbrev]{cleveref}

\theoremstyle{plain}

\theoremstyle{definition}

\theoremstyle{remark}

\usepackage[textsize=tiny]{todonotes}

\usepackage{soul, color}
\newcommand{\Note}[1]{}
\renewcommand{\Note}[1]{#1}  

\icmltitlerunning{Predicting Protein Variants}

\begin{document}

\twocolumn[
\icmltitle{Predicting protein variants with \\ equivariant graph neural networks}


\icmlsetsymbol{equal}{*}

\begin{icmlauthorlist}
\icmlauthor{Antonia Boca}{cam}
\icmlauthor{Simon V. Mathis}{cam}

\end{icmlauthorlist}

\icmlaffiliation{cam}{University of Cambridge}

\icmlcorrespondingauthor{Antonia Boca}{bocaantonia@gmail.com}
\icmlcorrespondingauthor{Simon Mathis}{simon.mathis@cl.cam.ac.uk}

\icmlkeywords{Machine Learning, ICML}

\vskip 0.3in
]



\printAffiliationsAndNotice{}

\begin{abstract}
Pre-trained models have been successful in many protein engineering tasks. 
Most notably, sequence-based models have achieved state-of-the-art performance on protein fitness prediction while structure-based models have been used experimentally to develop proteins with enhanced functions. 
However, there is a research gap in comparing structure- and sequence-based methods for predicting protein variants that are better than the wildtype protein. 
This paper aims to address this gap by conducting a comparative study between the abilities of equivariant graph neural networks (EGNNs) and sequence-based approaches to identify promising amino-acid mutations. 
The results show that our proposed structural approach achieves a competitive performance to sequence-based methods while being trained on significantly fewer molecules.
Additionally, we find that combining assay labelled data with structure pre-trained models yields similar trends as with sequence pre-trained models.

Our code and trained models can be found at: \url{https://github.com/semiluna/partIII-amino-acid-prediction}.

\end{abstract}

\section{Introduction}
\label{introduction}
In recent years, pre-trained models have garnered significant attention in the field of protein representation. Notably, models have been developed to deal with both the sequence and structure modalities of proteins \cite{ESM, prottrans, gearnet}. These models have demonstrated their potential in various applications such as protein fitness prediction \cite{ESM-1v, tranception} while being employed in a "zero-shot" manner, without the need for additional training data. Their success has also shown promising experimental results in protein engineering  \cite{mutcompute, Lu2022}. Additionally, \citet{chloe-hsu} have observed that augmenting simple models for assay labelled data with fitness predictions extracted from pre-trained sequence models can enhance their performance. 

Despite the experimental success of pre-trained structural methods for protein engineering, particularly those based on predicting residues given local atom environments \cite{torng20173d, Lu2022}, several crucial aspects remain unexplored. Firstly, these methods have not been systematically compared with sequence-based approaches using the same datasets. Secondly, their potential to augment assay labelled data, when available, has not been evaluated.

This paper aims to fill this research gap by conducting a study of the comparative performance of structure-based and sequence-based methods on predicting variants \textit{that are better than the wildtype protein}. 
We compare representatives of the most successful equivariant graph neural networks (EGNNs) on the task of residue identity prediction, namely GVP \cite{gvp2} and EQGAT \cite{eqgat}, with representatives of the most successful sequence-based approaches: Tranception \cite{tranception}, ESM-1v \cite{ESM-1v} and the MSA Transformer \cite{msa-transformer}.

By undertaking this comparative analysis, we aim to provide insights into the performance and suitability of geometric GNNs in protein engineering, specifically in the context of predictions based on the local atomic environment. Our contributions are as follows:
\begin{itemize}
    \item We apply the most successful pre-training approach for structural methods \cite{mutcompute} to equivariant GNNs by using the ATOM3D RES dataset \cite{atom3d} for residue identity prediction  (Table \ref{model-accuracy});
    \item We benchmark the resulting structure-based pre-trained models with the most successful zero-shot sequence-based approaches (Table \ref{generation-results}). We observe that structure does not trump sequence in downstream tasks when used in this way, although the amount of available structures used during pre-training is significantly lower than the number of sequences used in training large language models;
    \item We extend the simple combination approach for assay labelled data and pre-trained model outputs \cite{chloe-hsu} to the structure pre-trained domain. We find the same general trends as with sequence pre-trained models, as assay-labelled data quickly allows us to surpass zero-shot pre-trained sequence-based models with at few as 100 datapoints (Figure \ref{one-hot-regression}).
\end{itemize}

\section{Methodology}
We pre-train two equivariant graph neural networks on the task of residue identity prediction, also known as the RES task \cite{atom3d}.
We choose the Geometric Vector Perceptron \cite{gvp2} and the Equivariant Graph Attention Network \cite{eqgat}.
While \citet{Lu2022} used 3D-CNNs to engineer plastic enzymes, \citet{gvp2} benchmark 3D-CNNs on the RES task and show that the GVP outperforms them, so we choose to focus on this structural method instead.

Table \ref{model-accuracy} shows a comparison between the reported accuracies of the two models and the accuracies achieved in this paper. We achieve a higher performance on the GVP model than originally reported in \citet{gvp2}. This jump in performance can be explained by the fact that \citet{gvp2} only use a third of the original training dataset to train the GVP, possibly due to computational constraints. More details on our training parameters can be found in \ref{training-details}.

\begin{table}[t]
\caption{Classification accuracies on the ATOM3D RES dataset.}
\label{model-accuracy}
\vskip 0.15in
\begin{center}
\begin{small}
\begin{sc}
\begin{tabular}{@{}lcc@{}}
\toprule
Model & \begin{tabular}[c]{@{}c@{}}Reported \\ test accuracy\end{tabular} & \begin{tabular}[c]{@{}c@{}}Our \\ test accuracy\end{tabular} \\ \midrule
EQGAT & 0.540                                                             & 0.524                                                        \\
GVP   & 0.527                                                             & \textbf{0.580}                                               \\ \bottomrule
\end{tabular}
\end{sc}
\end{small}
\end{center}
\vskip -0.1in
\end{table}

\subsection{RES task formalism}
\label{res-task}
We formalise the RES classification task as follows. For a given point-cloud atomic graph $G = (V, E)$ with nodes $i,j \in V$ and edges $(i \rightarrow j) \in E$. Given a node $t \in V$ representing the $\text{C}_{\alpha}$ of a residue in the atomic graph, we can define the \textit{node classification function} $\text{RES}:\mathcal{V}\times\mathcal{G}\rightarrow\mathbb{R}^{20}$ that takes as input node $t$ and a \textit{masked} atomic graph $G_t$ from which we have removed the side-chain atoms of node $t$ and returns the likelihood scores of each of the 20 naturally occurring amino-acids to be part of the side-chain of node $t$. A more extended version of this formalism can be found in \ref{extended-res-formalism}.

\subsection{The scoring function}

We now formalise the function we use to score each amino-acid mutation in a sequence. For a wildtype protein sequence $x_1\dots x_n$ with $x_i\in\mathcal{A} = \{1,2,\dots,20\}$ we associate the point-cloud atomic graph $G = (V, E)$ corresponding to the protein's structure. Edges are drawn between any two atoms that are less than 4.5\AA ~apart.

Then, using the formalism defined in \ref{res-task}, the score associated with the presence of amino-acid $a\in\mathcal{A}$ at position $i$ can be defined as:
\begin{equation}
    S(i, a) = [\text{RES}(g(i), G)]_a
\label{scoring-function-short}
\end{equation}
Where $g:\{1,2,\dots,n\}\rightarrow\{1,2,\dots,|V|\}$ is a mapping function from positions to the index of the node representing the central $\text{C}_{\alpha}$ of the amino-acid residue present at each position. Here, $G_{g(i)}$ denotes the masked graph from which we removed the side-chain attached to node $g(i)$. 

Equation \ref{scoring-function-short} essentially represents the score of amino-acid $a$ for target position $i$, associated with node $g(i)$ in the atomic graph. Note that the true amino-acid at the same position is denoted by $x_i$.

\subsection{Mutation generation}
\begin{figure}[!h]
    \centering
    \includegraphics[width=\columnwidth]{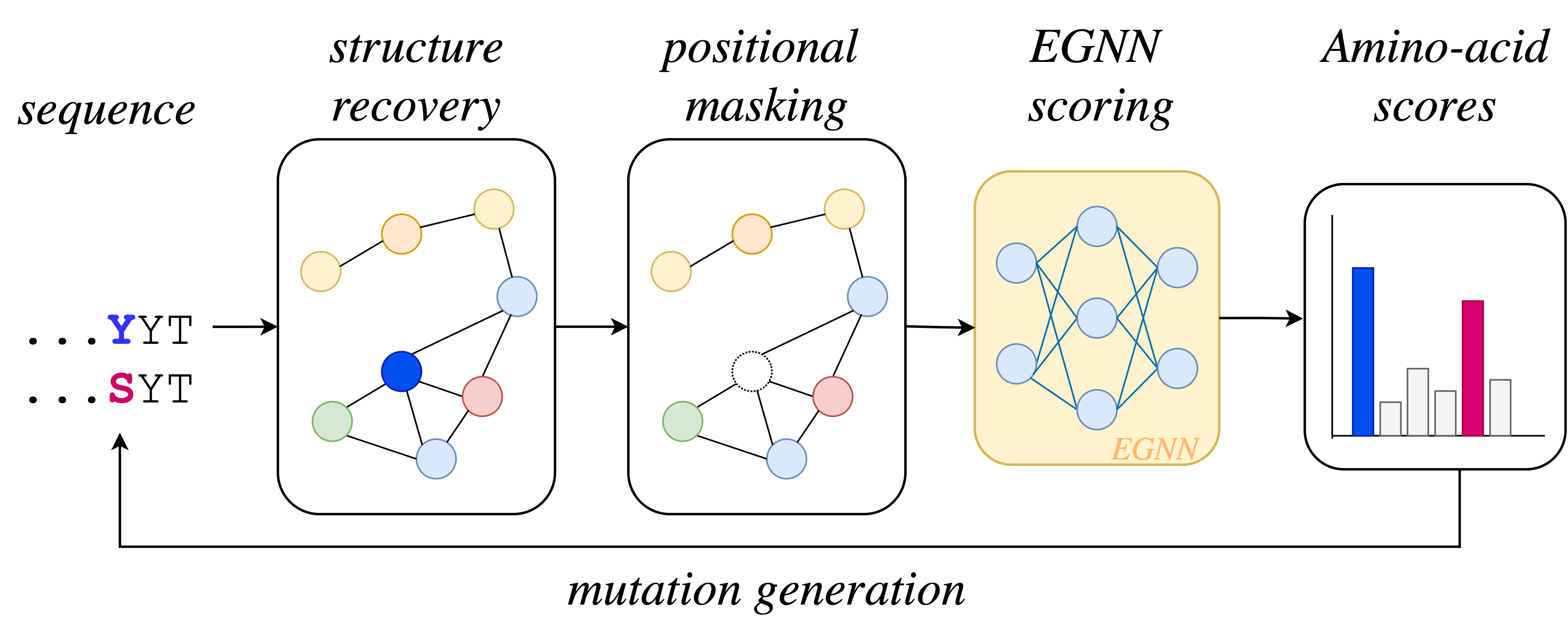}
    \caption{For every sequence, we recover the structure from the PDB and mask each amino-acid in turn. We pass the masked graph through a pre-trained EGNN model to recover the score associated with each amino-acid, which we then rank.  The key idea is that this pre-training allows the model to identify amino acids which seem “unusual” given their local environment and propose better fitting candidates instead.}
    \label{mutation-idea}
\end{figure}

Once the equivariant models have been trained on the RES task, we use them to inform the generation of single-point mutations in monomers and homo-oligomers from the ProteinGym substitutions dataset \cite{tranception}.
For each wildtype sequence we recover its structure, mask each amino-acid residue in turn, and retrieve the scores generated by the EGNN model for each of the 20 naturally occuring amino-acids. These scores are then ranked according to two strategies to determine the most promising single-point mutations. Figure \ref{mutation-idea} illustrates this approach visually.

\paragraph{Structure recovery.} 
The ProteinGym substitutions dataset contains 87 molecular sequences; for each of these sequences, a number of experimentally tested mutations are scored according to their \textit{fitness}. We evaluate our methods on a subset of the original dataset for which we could find either monomeric or homo-oligomeric structures. For each wildtype sequence, we recover the corresponding biological assembly from the Protein Data Bank \cite{rcsb_pdb}. When multiple assemblies are available, we choose one at random. When assemblies are incomplete, we instead use the monomeric AlphaFold prediction \cite{alpha_fold} if available. Otherwise, we discard the sequence. 

\subsection{Mutation ranking}
\label{sec:mutation-ranking}
Our approach allows us to score every possible residue mutation for each position in a sequence. Our goal is to generate meaningful mutations that have a higher chance of being bio-physically relevant, so we discard positions where the equivariant model makes the wrong prediction. A more detailed analysis of this design choice can be found in Appendix \ref{mutation-discard}.

\paragraph{Global ranking.} We rank the remaining mutations according to two strategies: \textit{global} and \textit{positional}. When performing global ranking, we sort mutations in descending order of their score, regardless of their position. If we denote the single-point mutation to amino-acid $a$ at position $i$ by $\mathbf{m}_{i}^a$, then $\forall i,j\text{ and }\forall a,b\in \mathcal{A} \text{ s.t. } a \neq x_i \text{ and } b \neq x_j$, we say that:
\begin{equation}
    \mathbf{m}_{i}^a\text{ is better than }\mathbf{m}_{j}^b \iff S(i, a) > S(j, b)
\label{global-ranking}
\end{equation}
\paragraph{Positional ranking.} 
The second approach follows when we prioritise the positions we want to mutate instead of the amino-acids we mutate to. Formally, this can be quantified as:
\begin{equation}
\begin{aligned}
&\mathbf{m}_{i}^a\text{ is better than }\mathbf{m}_{j}^b \\ 
\iff &\Big(S(i, x_i) < S(j, x_j)\Big) \lor \\ &\Big(S(i, x_i) = S(j, x_j) \land S(i, a) > S(j, b)\Big)
\end{aligned}
\label{positional-ranking}
\end{equation}
Note that when we perform positional ranking, we only keep the 3 top mutations for each position. 
\subsection{Protein fitness prediction}
\label{regression-methodology}
The GVP and EQGAT trained on the ATOM3D RES task can be thought of as unsupervised models that can suggest amino-acid mutations. We extend our original approach to perform fitness prediction using a ridge regression model augmented with the positional scores generated by equivariant GNNs, in a similar manner to that introduced by \citet{chloe-hsu}. For a given sequence of amino-acids $x_1\dots a_i \dots x_n$ with a single-point mutation at position $i$, we embed each amino-acid using either the one-hot encoding or \textit{AAIndex} embeddings \cite{aa-index} on which we perform PCA to render 19-dimensional features per amino-acid. We flatten and concatenate these encodings to render feature vectors $\mathbf{h}_{\text{one-hot}}\in\mathbb{R}^{20\times n}$ and $\mathbf{h}_{\text{aa-index}}\in\mathbb{R}^{19\times n}$. To this feature vector we concatenate the score predicted by the GNN model for amino-acid $a_i$ at position $i$:
\begin{align}
    \mathbf{x}_{\text{one-hot}} &= [\mathbf{h}_{\text{one-hot}} ~||~ S(i, a_i)] \\
    \mathbf{x}_{\text{aa-index}} &= [\mathbf{h}_{\text{aa-index}} ~||~ S(i, a_i)]
\end{align}
Here, $S(i, a_i)$ is the same scoring function defined in Equation \ref{scoring-function-short}.
These features are then used to train a ridge regression model to predict protein fitness using subsets of single-point mutated sequences for each of the ProteinGym DMS assays we have model scores for.

\begin{table*}[tb]
\caption{\raggedright{Ranking performance of the models across 49 DMS assays. Numbers in \textbf{bold} represent the highest score per column, while numbers with an \underline{underline} represent the second highest score per column. We note that two equivariant GNNs have the highest rank correlation for better than wildtype mutations.}}
\label{generation-results}
\vskip 0.15in
\begin{center}
\begin{small}
\begin{sc}
\begin{tabular}{@{}ccccccc@{}}
\toprule
\multirow{2}{*}{Model} & \multirow{2}{*}{Ranking strategy} & \multirow{2}{*}{\begin{tabular}[c]{@{}c@{}}Top 10\\ precision\end{tabular}} & \multirow{2}{*}{\begin{tabular}[c]{@{}c@{}}Top 10\\ recall\end{tabular}} & \multicolumn{3}{c}{Spearman's rank correlation}  \\ \cmidrule(l){5-7} 
                       &                                   &                                                                             &                                                                          & Average        & Worse than WT  & Better than WT \\ \midrule
EQGAT                  & Positional                        & 0.486                                                                       & \underline{0.187}                                                              & 0.223          & 0.128          & 0.118          \\
EQGAT                  & Global                            & 0.491                                                                       & 0.072                                                                    & 0.262          & 0.154          & \underline{0.157}    \\
GVP                    & Positional                        & 0.462                                                                       & \textbf{0.419}                                                           & 0.106          & $-0.009$         & \textbf{0.276} \\
GVP                    & Global                            & 0.426                                                                       & 0.100                                                                    & 0.202          & 0.128          & $-0.011$         \\ \midrule
Tranception            &                                   & \underline{0.619}                                                                 & 0.012                                                                    & \underline{0.429}    & \underline{0.299}    & 0.143          \\
ESM-1v                &                                   & 0.618                                                                       & 0.018                                                                    & 0.407          & 0.288          & 0.135          \\
MSA Transformer        &                                   & \textbf{0.638}                                                              & 0.018                                                                    & \textbf{0.434} & \textbf{0.327} & 0.135          \\ \bottomrule
\end{tabular}
\end{sc}
\end{small}
\end{center}
\vskip -0.1in
\end{table*}

\section{Results}
\subsection{Mutation generation}
We generate single-point mutations for 49 out of the 87 DMS assays in the ProteinGym substitutions dataset \cite{tranception}. When we generate mutations, we discard any that we cannot find in the experimental dataset of the target sequence. We are interested in understanding how good our models are at suggesting mutations that are \textit{better than the wildtype} sequence, hence we propose three metrics through which to perform comparisons: (1) Spearman's rank correlation restricted to better than wildtype sequences, (2) the precision of the top 10 mutations, and (3) the recall of the top 10 mutations. To compute the last two metrics we only considered whether a mutation proposed by the model is better than the wildtype, disregarding its actual score. Table \ref{generation-results} shows the performance of our models, depending on the type of ranking used. We note that the equivariant models have a competitive performance to Tranception \cite{tranception} when ranking mutations that are better than the wildtype, indicating that they represent a viable strategy for aiding the discovery process in protein engineering. 
Per-dataset performance metrics can be found in \ref{per-dataset-performance}.

EGNN models require a significantly smaller number of protein structures during training in order to reach a similar ranking correlation coefficient to Tranception for mutations that are better than the wildtype. While Tranception is trained on the UniRef100 database \cite{uniref}, which contains over 4 million source sequences, our models are trained on the ATOM3D RES dataset \cite{atom3d}, which contains fewer than 22k molecules from which local environments are sampled. 

\paragraph{Structure vs Sequence.} We believe EGNNs may require less training because structure is more informative than sequence for fitness prediction. While sequence-based models attend the full protein and subsequently learn to focus on the important bits, EGNNs attend only local environments, thus learning to identify important atoms faster. Further experiments could be run to compare the power of sequence- and structure-based models when their level of training is comparable. However, we point out that training EGNNs to the same level as present state-of-the-art sequence models may be infeasible due to both data and computational constraints.

\paragraph{Correlation to sequence-based models.} As part of our analysis, we also compute the correlation between the better than wildtype predictions made by our EGNN models and Tranception. Per-model and per-dataset statistics can be found in \ref{tranception-correlation}; we note that the highest rank correlation we find is \textbf{0.212}, in the case of the EQGAT model. Since these approaches seem to be weakly correlated, we believe there are improvements to be gained from ensembling both structure- and sequence-based approaches.

\paragraph{The impact of design choices.} As mentioned in Section \ref{sec:mutation-ranking}, we discard mutations at positions where the EGNNs make the wrong prediction, as we find that incorporating these is detrimental to the overall performance (see \ref{mutation-discard}). This indicates that these structure-based models are still undertrained, with potential for improvement coming both from larger datasets and more data engineering.
\subsection{Protein fitness prediction}
\begin{figure}[!htb]
    \centering
    \includegraphics[width=\columnwidth]{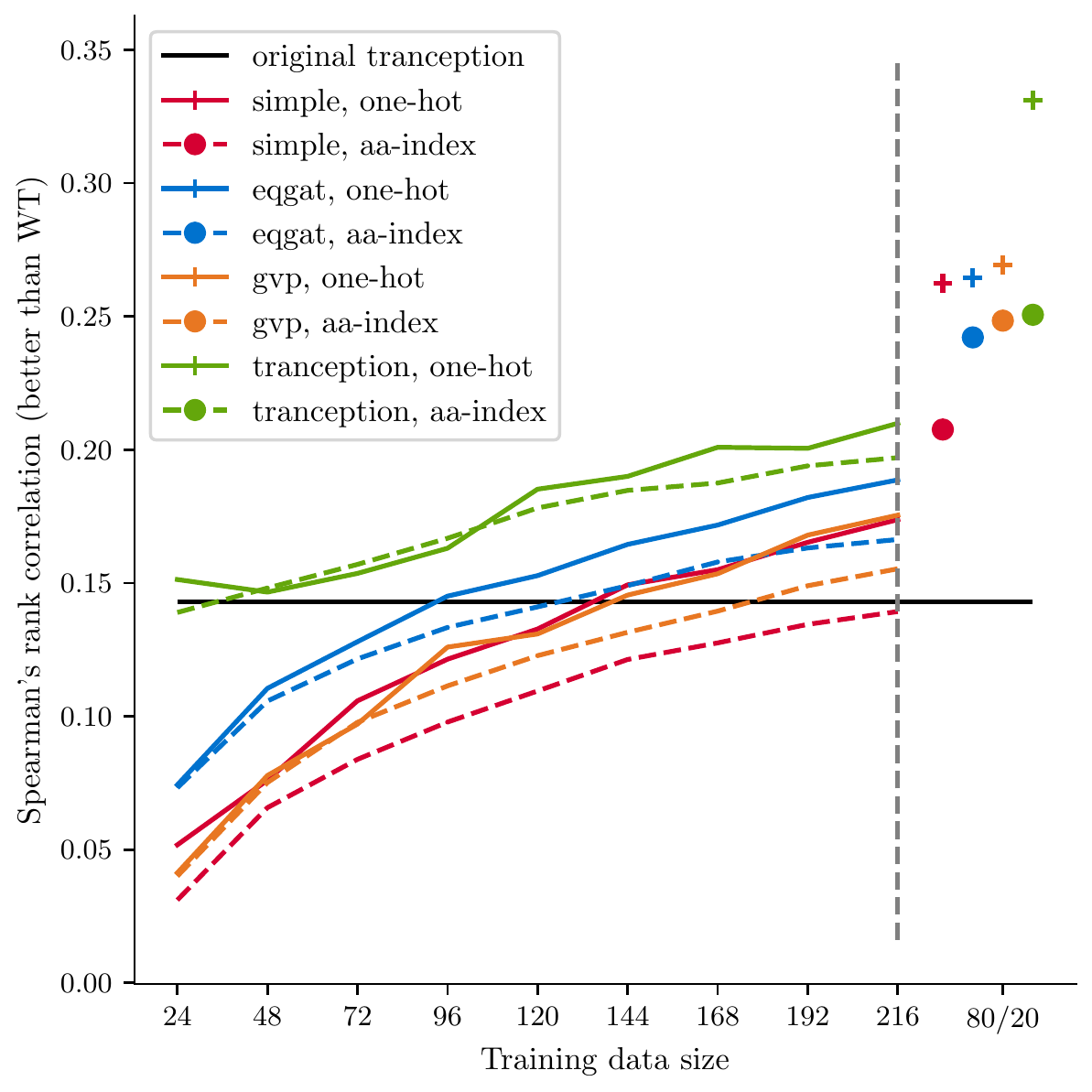}
    \caption{\raggedright Performance on mutations that are better than the wildtype for four regression models using two types of embeddings. Statistics are aggregated across 49 DMS assays. 
    We note that we can improve the fitness prediction performance above the Tranception baseline (in black) across all regression models by training on as few as 144 data points. 
    }
    \label{one-hot-regression}
\end{figure}

We train 4 types of ridge regression models on each of the 49 DMS datasets separately. The baseline non-augmented model uses only features $\mathbf{h}_{\text{one-hot}}$ or $\mathbf{h}_{\text{aa-index}}$ defined in Section \ref{regression-methodology}; the remaining 3 models are augmented with single-point mutation scores from GVP, EQGAT, and Tranception, respectively. For each model type and each DMS array we first set aside 20\% of the single-point mutated sequences for testing; 
We train the regression on increasingly larger training subsets. We repeat the process 20 times with different random subsets and report the average Spearman rank correlation on better than wildtype sequences, as seen in  Figure \ref{one-hot-regression}. The performance on other metrics can be found in \ref{ridge-regression-all-stats}.

Similar to the results reported by \citet{chloe-hsu}, the augmented linear models allow us to surpass the baseline zero-shot fitness prediction models with as few as 100 datapoints in the case of the model augmented with EQGAT scores. While the linear model augmneted with Tranception scores performs best overall, we point out that Tranception is fine-tuned to predict protein fitness, while the scores retrieved from our models merely represent the confidence in a certain amino-acid for a target position.

\section{Limitations and future work}
We apply pre-trained EGNNs to both mutation generation and protein fitness prediction, and find that structural approaches are a competitive approach to sequence-based language models for the prediction of mutations that are better than the wildtype, while also requiring \textbf{181x} fewer molecules to train. 

While the results look promising, this comparison is limited in scope, as our approach does not deal with more complex (hetero-oligomeric) structures from the ProteinGym dataset. 

\paragraph{Types of fitness.} Additionally, the benchmarking dataset contains a wide range of sequences for which ``fitness'' can be interpreted in many different ways. 
DMS assays in the ProteinGym dataset come from humans, viruses, prokaryotes, and eukaryotes. In particular, in the subset of 49 DMS assays used in this paper, \textbf{5} come from eukaryotes, \textbf{21} from humans, \textbf{18} from prokaryotes, and \textbf{5} from viruses. 
Fitness, in the case of viruses, is interpreted as infectivity or the likelihood of mutation. In the rest of the cases, fitness can range from stress resistance to efficiency. For example, in their experimental paper, \citet{Lu2022} focused on improving thermal stability. Hence, the fitness score used by \citet{tranception} in the ProteinGym dataset represents a ``fuzzy'' concept that is context-dependent.
Future work could focus more closely on identifying the types of fitness structure-based approaches excel at. 


\paragraph{Acknowledgements} SVM was supported by the UKRI Centre for Doctoral Training in
Application of Artificial Intelligence to the study of Environmental Risks (EP/S022961/1).

\FloatBarrier
\bibliography{bibliography}
\bibliographystyle{styles/icml2023}

\newpage
\appendix
\onecolumn
\section{Appendix}
\subsection{Extended RES task formalism}
\label{extended-res-formalism}

For a given point-cloud atomic graph $\mathcal{G} = (\mathcal{V}, \mathcal{E})$ with nodes $i,j \in \mathcal{V}$ and edges $(i \rightarrow j) \in \mathcal{E}$ for which we have initial scalar and vector node features $\mathbf{H}\in\mathbb{R}^{|\mathcal{V}|\times n}\times\mathbb{R}^{|\mathcal{V}|\times 3 \times \nu}$, as well as scalar and vector edge features $\mathbf{E}\in\mathbb{R}^{|\mathcal{E}|\times m}\times\mathbb{R}^{|\mathcal{E}|\times 3 \times \eta}$, we first consider a \textit{masked} version of this graph, $\mathcal{G}_t=(\mathcal{V}_t, \mathcal{E}_t)$, for which we have masked all the atoms of the side-chain attached to the node $t$ (representing an $\text{C}_{\alpha}$ atom). 

These features become the input to an EGNN model trained on the RES task, that returns the probability that the masked residue $t$ is any of the 20 naturally occurring amino-acids. Formally, if we define the \textit{node classification function} $f_{\gamma}^t:\mathbb{R}^{|\mathcal{V}|\times n}\times\mathbb{R}^{|\mathcal{V}|\times 3 \times \nu}\times\mathbb{R}^{|\mathcal{E}|\times m}\times\mathbb{R}^{|\mathcal{E}|\times 3 \times \eta}\rightarrow \mathbb{R}^{20}$ with learnable parameters $\gamma$, then the score of amino-acid $a$ at position $i$ in a wildtype sequence can be defined as:
\begin{equation}
    S(i, a) = [f_{\gamma}^{g(i)}(\mathbf{H}, \mathbf{E})]_a
\label{scoring-function}
\end{equation}
We then formally define the GNN model $\mathbf{G}_{\theta_1}:\mathbb{R}^{|\mathcal{V}_t|\times n}\times\mathbb{R}^{|\mathcal{V}_t|\times 3 \times \nu}\times\mathbb{R}^{|\mathcal{E}_t|\times m}\times\mathbb{R}^{|\mathcal{E}_t|\times 3 \times \eta}\rightarrow \mathbb{R}^{|\mathcal{V}_t|\times o}$ that takes as input the node and edge features and returns final node features  $\mathbf{H}_{\text{out}}$:
\begin{equation}
    \mathbf{H}_{\text{out}} = \mathbf{G}_{\theta_1}(\mathbf{H}_t, \mathbf{E}_t)
\end{equation}
where $\mathbf{H}_t$ and $\mathbf{E}_t$ represent the node and edge features for all nodes and edges that exist in the masked graph $\mathcal{G}_t$.

Since we are interested in predicting the type of amino-acid corresponding to the masked side-chain of node $t$, we pass its final features $[\mathbf{H}_{\text{out}}]_{t}$ through a multi-layer perceptron $\text{MLP}_{\theta_2}:\mathbb{R}^o\rightarrow \mathbb{R}^{20}$ to obtain the final scores associated to each of the 20 naturally occuring amino-acids.

\begin{equation}
    f_{\gamma}^t(\mathbf{H}, \mathbf{E}) = \text{MLP}_{\theta_2}([\mathbf{G}_{\theta_1}(\mathbf{H}_t, \mathbf{E}_t)]_t)
\label{full-formalism}
\end{equation}

Given a wildtype sequence $x_1x_2\dots x_n$ of length $n$ with $x_i \in \mathcal{A} = \{1,2,\dots,20\}$ representing the index of amino-acid $i$, we construct the atomic graph $\mathcal{G} = (\mathcal{V}, \mathcal{E})$, and build a scoring function of the positions  $S:\{1,2,\dots,n\}\times\mathcal{A}\rightarrow \mathbb{R}$ that we define by extending the formalism in Equation \ref{full-formalism}:
\begin{equation}
    S(i, a) = [f_{\gamma}^{g(i)}(\mathbf{H}, \mathbf{E})]_a
\label{scoring-function}
\end{equation}
Where $g:\{1,2,\dots,n\}\rightarrow\{1,2,\dots,|\mathcal{V}|\}$ is a mapping function from positions to the index of the node representing the central $\text{C}_{\alpha}$ of the residue at every position in the sequence.

\newpage
\subsection{Per-dataset performance of our ranking approach}
\label{per-dataset-performance}
\begin{figure}[!h]
    \centering
    \includegraphics[width=\textwidth]{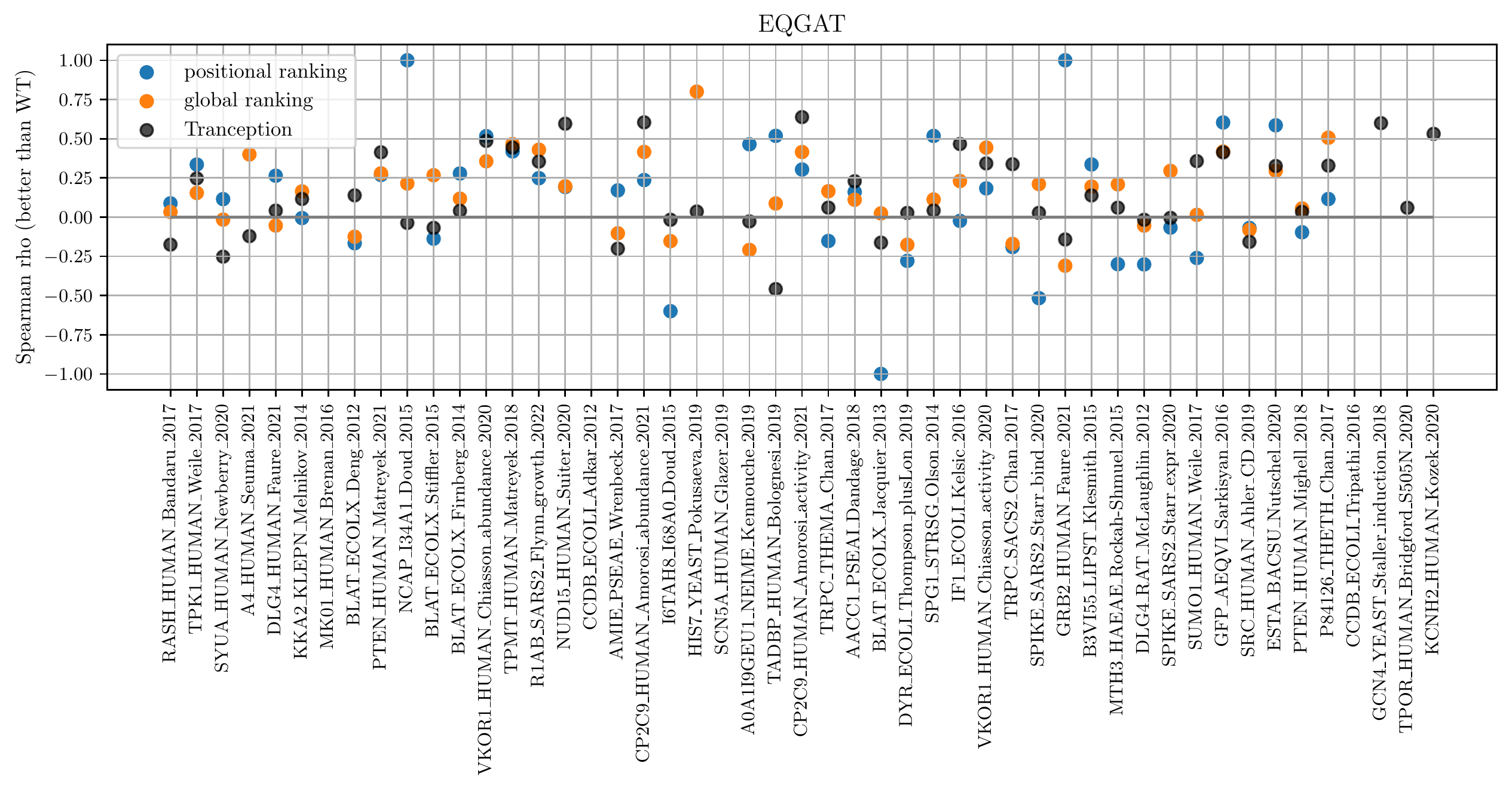}
    \caption{Spearman's rank correlation on better than wildtype mutations per dataset for the EQGAT model.}
    \label{fig:eqgat-per-dataset}
\end{figure}

\begin{figure}[!h]
    \centering
    \includegraphics[width=\textwidth]{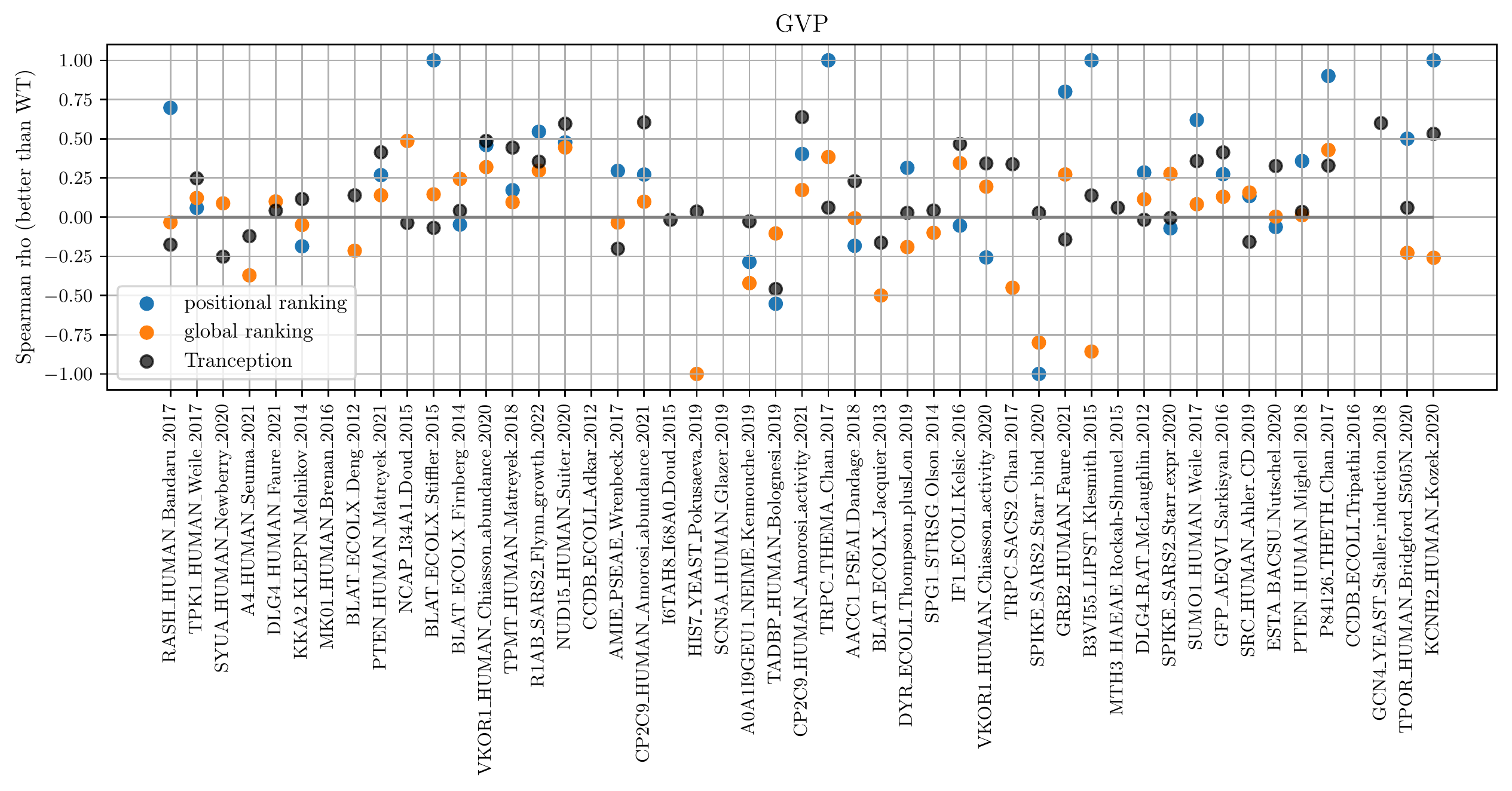}
    \caption{Spearman's rank correlation on better than wildtype mutations per dataset for the GVP model.}
    \label{fig:gvp-per-dataset}
\end{figure}
\FloatBarrier

\newpage
\subsection{Correlation to Tranception ranking}
\label{tranception-correlation}

\begin{table}[!h]
\centering
\caption{Average Spearman rank correlation (for better than wildtype predictions) between Tranception and structure-based models. }
\vskip 0.15in
\begin{tabular}{@{}ccc@{}}
\toprule
Model                  & Ranking    & Correlation to Tranception \\ \midrule
\multirow{2}{*}{EQGAT} & Global     & 0.212                      \\
                       & Positional & 0.146                      \\ \midrule
\multirow{2}{*}{GVP}   & Global     & 0.147                      \\
                       & Positional & 0.073                      \\ \bottomrule
\end{tabular}

\end{table}

\begin{figure}[!h]
    \centering
    \includegraphics[width=\textwidth]{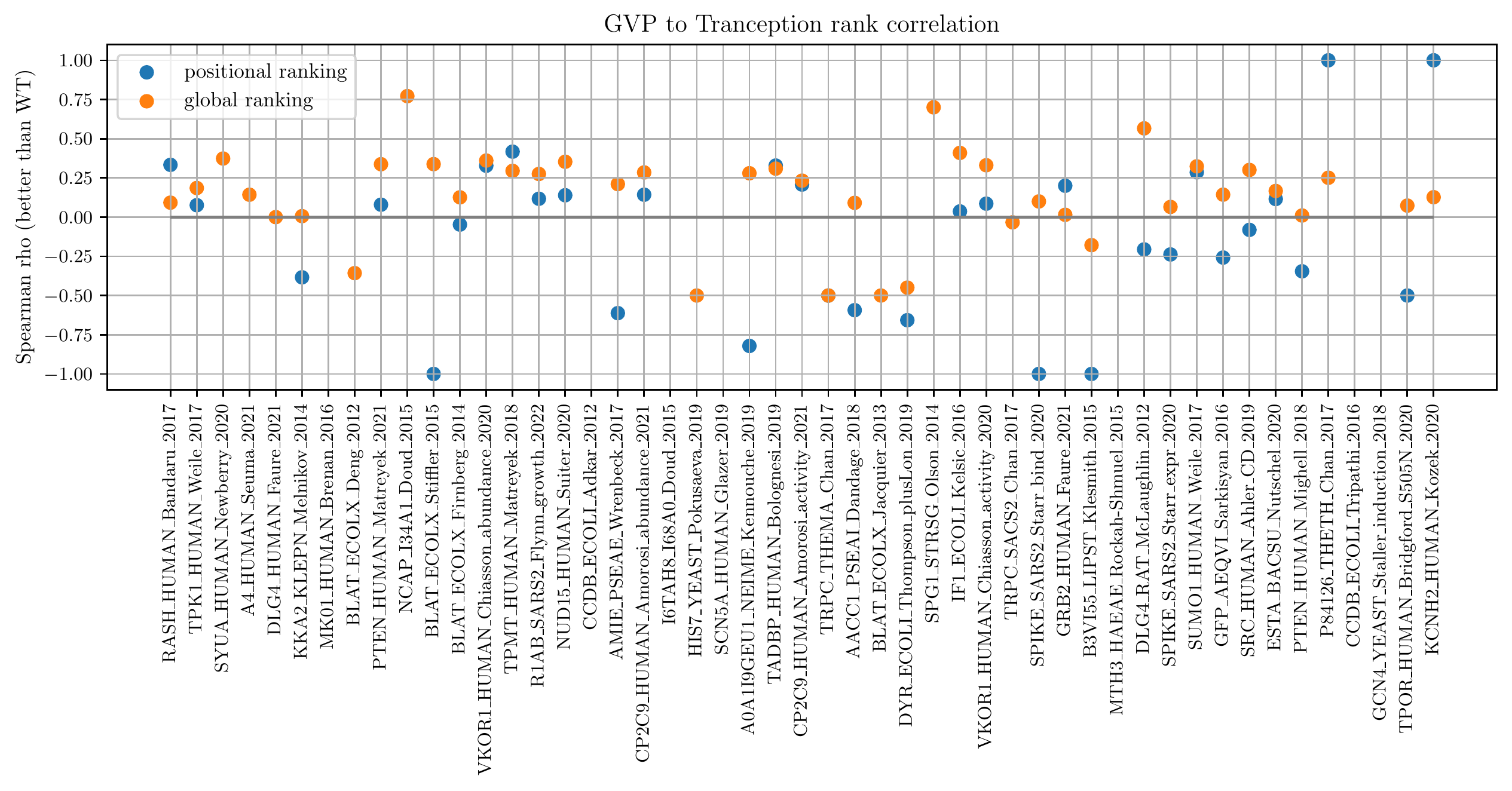}
    \caption{Spearman's rank correlation between the ranking made by the GVP model and Tranception.}
    \label{fig:gvp-tranception-corr}
\end{figure}
\begin{figure}[!h]
    \centering
    \includegraphics[width=\textwidth]{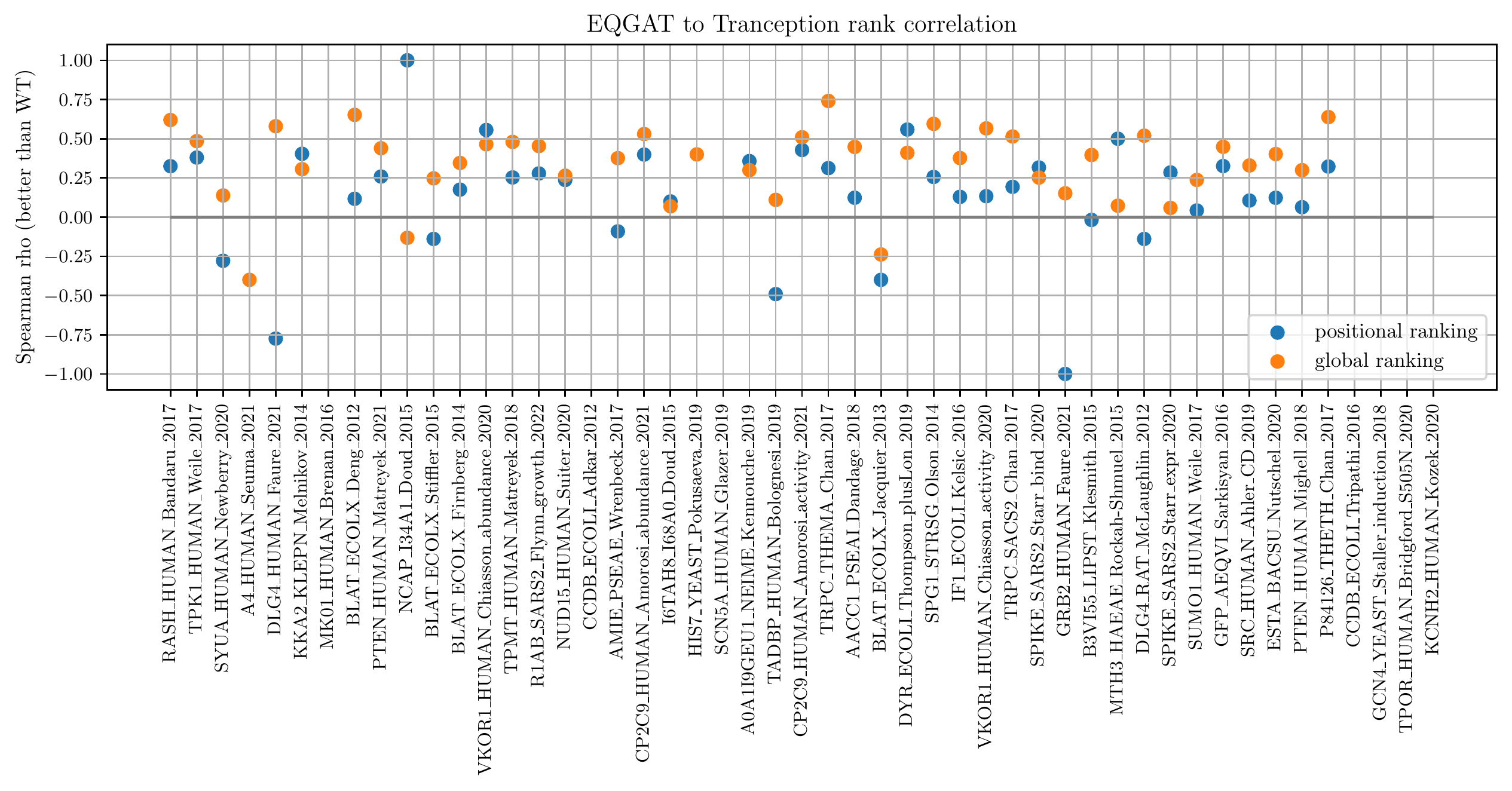}
    \caption{Spearman's rank correlation between the ranking made by the EQGAT model and Tranception.}
    \label{fig:eqgat-tranception-corr}
\end{figure}

\FloatBarrier
\subsection{Design choice: discarding wrongly predicted positions}
\label{mutation-discard}

As mentioned at the beginning of Section \ref{sec:mutation-ranking}, we add an additional filter to our ranking techniques in order to increase the quality of the mutations proposed by our models. More specifically, we discard the scores of all mutations at a position that was classified incorrectly by the EGNN. 

The reasoning behind this is that an incorrect classification of a residue may indicate that the models do not have a good understanding of the biophysical properties of the residue's local environment. This is particularly damaging to the generation of meaningful mutations if the models assign similarly low confidences to all 20 amino-acids that are candidate for a position, because they resort to random guessing. 

We provide an ablation study to back up this design choice. Tables \ref{ablation-correct-only-global} and \ref{ablation-correct-only-positional} show how all three types of Spearman rank correlations are higher when discarding wrongly predicted positions. 

\begin{table}[!h]
\caption{Model performance when performing global ranking using both wrongly and correctly predicted positions. Statistics averaged across for 49 sequences.}
\label{ablation-correct-only-global}
\vskip 0.15in
\begin{center}
\begin{small}
\begin{sc}
\begin{tabular}{@{}ccccc@{}}
\toprule
\multirow{2}{*}{Model} & \multirow{2}{*}{Positions used} & \multicolumn{3}{c}{Spearman's rank correlation}  \\ \cmidrule(l){3-5} 
                       &                                 & Average        & Worse than WT  & Better than WT \\ \midrule
EQGAT                  & all                             & 0.20           & 0.153          & 0.069          \\
EQGAT                  & correct only                    & \textbf{0.262} & \textbf{0.154} & \textbf{0.157} \\ \midrule
GVP                    & all                             & 0.093          & 0.076          & \textbf{0.014} \\
GVP                    & correct only                    & \textbf{0.202} & \textbf{0.128} & -0.011         \\ \midrule
Tranception            &                                 & 0.429          & 0.299          & 0.143          \\ \bottomrule
\end{tabular}
\end{sc}
\end{small}
\end{center}
\vskip -0.1in
\end{table}

\begin{table}[!h]
\caption{Model performance when performing positional ranking using both wrongly and correctly predicted positions. Statistics averaged across for 49 sequences.}
\label{ablation-correct-only-positional}
\vskip 0.15in
\begin{center}
\begin{small}
\begin{sc}
\begin{tabular}{@{}ccccc@{}}
\toprule
\multirow{2}{*}{Model} & \multirow{2}{*}{Positions used} & \multicolumn{3}{c}{Spearman's rank correlation}             \\ \cmidrule(l){3-5} 
                       &                                 & Average        & Worse than WT & Better than WT \\ \midrule
EQGAT                  & all                             & 0.124          & 0.100               & 0.061                \\
EQGAT                  & correct only                    & \textbf{0.223} & \textbf{0.128}      & \textbf{0.118}       \\ \midrule
GVP                    & all                             & 0.083          & \textbf{0.076}      & -0.018               \\
GVP                    & correct only                    & \textbf{0.106} & -0.009              & \textbf{0.276}       \\ \midrule
Tranception            &                                 & 0.429          & 0.299               & 0.143                \\ \bottomrule
\end{tabular}
\end{sc}
\end{small}
\end{center}
\vskip -0.1in
\end{table}

\begin{figure}
    \centering
    \includegraphics[width=\textwidth]{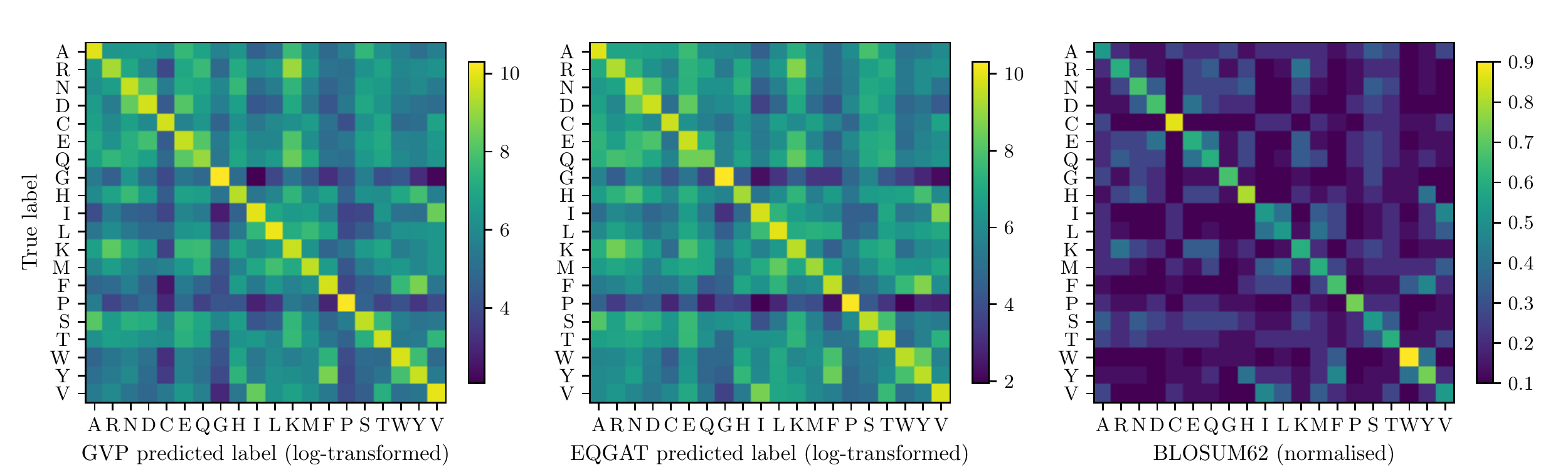}
    \caption{Comparison between the confusion matrices of EQGAT and GVP to the BLOSUM62 matrix.}
    \label{fig:confusions}
\end{figure}

We can get a better idea of the types of information that is learned by the EGNN models by looking at the confusion matrices between the true labels and the predicted labels in the RES ATOM3D test dataset, illustrated in Figure \ref{fig:confusions}. 

\paragraph{The BLOSUM62 matrix.} We compare these two matrices to the BLOSUM62 matrix, a scoring matrix commonly used in bioinformatics for sequence alignment. It stands for "BLOcks SUbstitution Matrix" and was developed by \citet{Henikoff1992}. In the BLOSUM62 matrix, each cell represents the score for substituting one amino acid with another. The matrix is symmetric, and the scores are typically positive integers. Higher scores indicate a higher degree of conservation or similarity between the substituted amino acids.

We expect that model with meaningful inferred knowledge about the biophysical properties of molecules and amino-acids to have a similar confusion matrix to the BLOSUM62 matrix, yet we do not necessarily see this trend arising. We do, however, notice that the confusion matrices of the EQGAT and the GVP look similar to each other, indicating that they must be learning the same features from the training dataset.  

\subsection{Design choice: experimental vs. AlphaFold structures}
\label{alpha-fold-comparison}

We expect the usage of AlphaFold structures to be detrimental to the overall performance of our models. To check whether this is indeed the case, we perform on ablation study on 14 of the 49 sequences we use in this project. These sequences have both a complete experimental structure and an AlphaFold structure, so we can compare the performance of our models using either one or the other. Since the ablation study presented in Section \ref{mutation-discard} makes it clear that mutations at wrongly predicted positions are detrimental to our models, we perform this second ablation study by also discarding mutations for positions that models get wrong.

As presented in Tables \ref{ablation-alphafold-global} and \ref{ablation-alphafold-positional}, we find that there isn't a clear relation between using the AlphaFold structure and a decrease in the ``better than WT'' Spearman correlation, as it seeems to depend on both the model and the ranking strategy used. However, we \textit{do} notice a clear trend for the models rank \textit{worse than wildtype} mutations better when using the AlphaFold structure. 
\begin{table}[!h]
\caption{Model performance when performing global ranking using either AlphaFold or experimental features. Statistics are averages across 14 DMS assays.}
\label{ablation-alphafold-global}
\vskip 0.15in
\begin{center}
\begin{small}
\begin{sc}
\begin{tabular}{@{}ccccc@{}}
\toprule
\multirow{2}{*}{Model} & \multirow{2}{*}{Structure} & \multicolumn{3}{c}{Spearman's rank correlation}  \\ \cmidrule(l){3-5} 
                       &                            & Average        & Worse than WT  & Better than WT \\ \midrule
EQGAT                  & AlphaFold                  & \textbf{0.311} & \textbf{0.177} & 0.136          \\
EQGAT                  & Experimental               & 0.262          & 0.154          & \textbf{0.157} \\ \midrule
GVP                    & AlphaFold                  & \textbf{0.237} & \textbf{0.211} & \textbf{0.049} \\
GVP                    & Experimental               & 0.202          & 0.128          & $-0.011$         \\ \bottomrule
\end{tabular}
\end{sc}
\end{small}
\end{center}
\vskip -0.1in
\end{table}

\begin{table}[!h]
\caption{Model performance when performing positional ranking using either AlphaFold or experimental features. Statistics are averages across 14 DMS assays.}
\label{ablation-alphafold-positional}
\vskip 0.15in
\begin{center}
\begin{small}
\begin{sc}

\begin{tabular}{@{}ccccc@{}}
\toprule
\multirow{2}{*}{Model} & \multirow{2}{*}{Structure} & \multicolumn{3}{c}{Spearman's rank correlation}  \\ \cmidrule(l){3-5} 
                       &                            & Average        & Worse than WT  & Better than WT \\ \midrule
EQGAT                  & AlphaFold                  & \textbf{0.235} & 0.097          & \textbf{0.149} \\
EQGAT                  & Experimental               & 0.223          & \textbf{0.128} & 0.118          \\ \midrule
GVP                    & AlphaFold                  & \textbf{0.253} & \textbf{0.332} & 0.172          \\
GVP                    & Experimental               & 0.106          & -0.009         & \textbf{0.276} \\ \bottomrule
\end{tabular}

\end{sc}
\end{small}
\end{center}
\vskip -0.1in
\end{table}

\subsection{Design choice: full structure vs. local environment}

 For the main results, we input the entire molecule in the GNN (with the exception of the masked amino-acid position we wish to predict scores for). This may not necessarily be the best approach, because the models are trained on \textit{samples of local environments}, which contain on average 600 nodes, whereas a full molecular structure can have even 4000 nodes. Tables \ref{ablation-local-environment-positional} and \ref{ablation-local-environment-global} show that the EQGAT model benefits from using the entire structure, while the GVP model benefits from using the local environment. 
 
\begin{table}[!h]
\caption{Model performance when performing positional ranking using either local environments or the full molecule. Statistics are averaged across 49 DMS assays.}

\label{ablation-local-environment-positional}
\vskip 0.15in
\begin{center}
\begin{footnotesize}
\begin{sc}
\begin{tabular}{@{}ccccccc@{}}
\toprule
\multirow{2}{*}{Model} & \multirow{2}{*}{Structure} & \multicolumn{3}{c}{Spearman's rank correlation}  & \multirow{2}{*}{\begin{tabular}[c]{@{}c@{}}Top 10 \\ precision\end{tabular}} & \multirow{2}{*}{\begin{tabular}[c]{@{}c@{}}Top 10 \\ recall\end{tabular}} \\ \cmidrule(lr){3-5}
                       &                            & Average        & Worse than WT  & Better than WT &                                                                              &                                                                           \\ \midrule
EQGAT                  & Full                       & \textbf{0.223} & \textbf{0.128} & \textbf{0.118} & 0.486                                                                        & \textbf{0.187}                                                            \\
EQGAT                  & Local                      & 0.203          & 0.039          & 0.041          & \textbf{0.516}                                                               & 0.176                                                                     \\ \midrule
GVP                    & Full                       & 0.106          & -0.009         & 0.276          & \textbf{0.462}                                                               & \textbf{0.419}                                                            \\
GVP                    & Local                      & \textbf{0.203} & \textbf{0.104} & \textbf{0.311} & 0.451                                                                        & 0.382                                                                     \\ \bottomrule
\end{tabular}
\end{sc}
\end{footnotesize}
\end{center}
\vskip -0.1in
\end{table}

\begin{table}[!h]
\caption{Model performance when performing global ranking using either local environments or the full molecule. Statistics are averaged across 49 DMS assays.}

\label{ablation-local-environment-global}
\vskip 0.15in
\begin{center}
\begin{footnotesize}
\begin{sc}
\begin{tabular}{@{}ccccccc@{}}
\toprule
\multirow{2}{*}{Model} & \multirow{2}{*}{Structure} & \multicolumn{3}{c}{Spearman's rank correlation}   & \multirow{2}{*}{\begin{tabular}[c]{@{}c@{}}Top 10 \\ precision\end{tabular}} & \multirow{2}{*}{\begin{tabular}[c]{@{}c@{}}Top 10 \\ recall\end{tabular}} \\ \cmidrule(lr){3-5}
                       &                            & Average        & Worse than WT  & Better than WT  &                                                                              &                                                                           \\ \midrule
EQGAT                  & Full                       & \textbf{0.262} & \textbf{0.154} & \textbf{0.157}  & 0.491                                                                        & \textbf{0.072}                                                            \\
EQGAT                  & Local                      & 0.254          & 0.149          & 0.134           & \textbf{0.491}                                                               & 0.047                                                                     \\ \midrule
GVP                    & Full                       & 0.202          & 0.128          & \textbf{-0.011} & \textbf{0.426}                                                               & 0.100                                                                     \\
GVP                    & Local                      & \textbf{0.216} & 0.233          & \textbf{-0.031} & 0.392                                                                        & \textbf{0.126}                                                            \\ \bottomrule
\end{tabular}
\end{sc}
\end{footnotesize}
\end{center}
\vskip -0.1in
\end{table}
\newpage
\subsection{Summary of design decisions for mutation generation}
Overall, our ablation studies suggest the following:
\begin{itemize}
    \item Discarding positions where the models cannot correctly identify the true wildtype amino-acid increases the quality the mutation ranking;
    \item Using the AlphaFold structure is \textit{not} detrimental to the generation of better than wildtype mutations;
    \item The EQGAT model works best with the full molecular structure, while the GVP model works best with the a local environment.
\end{itemize}
\FloatBarrier
\subsection{Performance metrics for the ridge regression models}
\label{ridge-regression-all-stats}
\begin{figure}[!h]
    \centering
    \subfigure[\raggedright{Average Spearman rank correlation for all ridge regression models.}]{
        \includegraphics[scale=0.5]{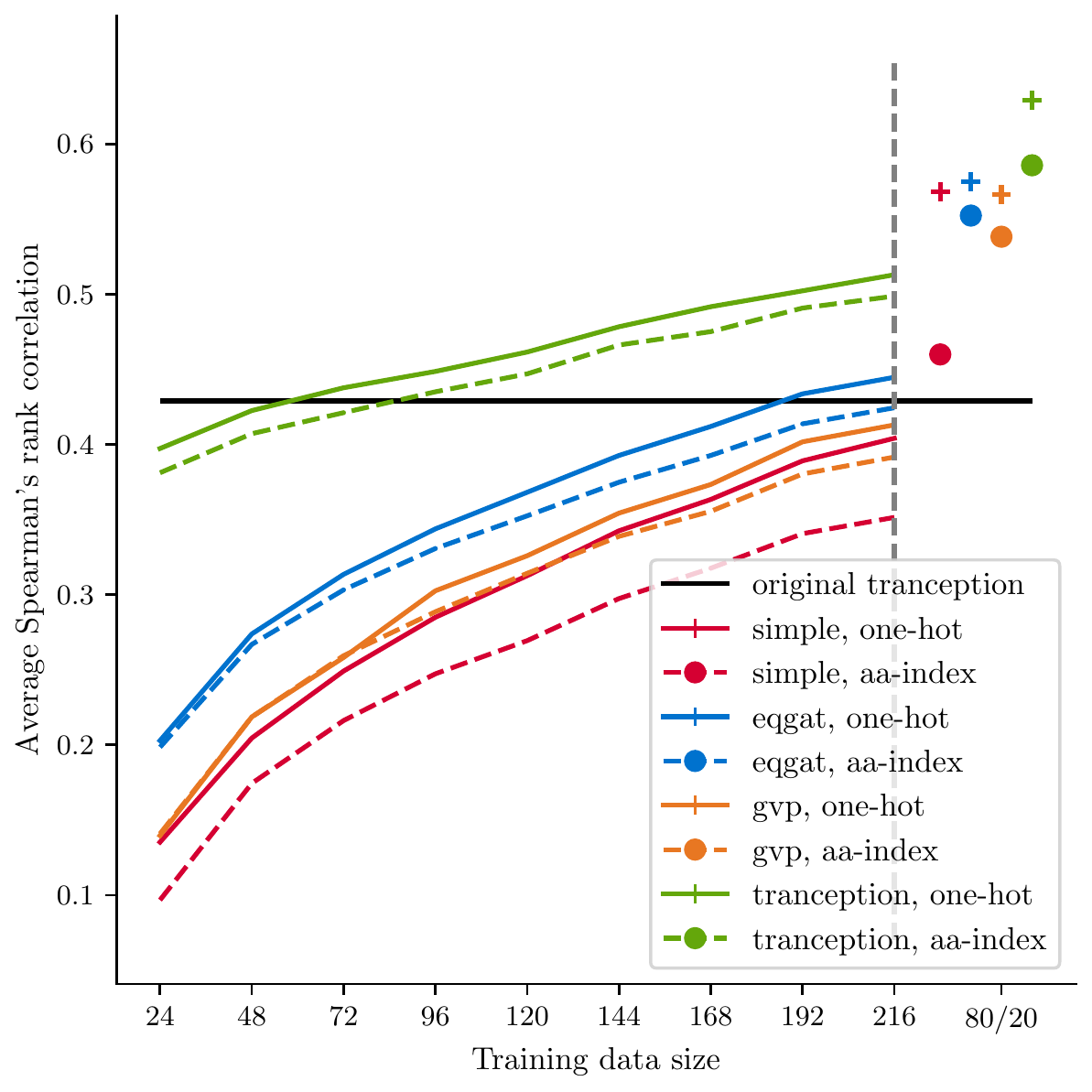}
    }
    \hspace{0.3in}
    \subfigure[\raggedright{Average Spearman rank correlation (better than WT) for all ridge regression models.}]{
        \includegraphics[scale=0.5]{figs/better_than_WT_spearmanr_all_square.pdf}
    }

    \vspace{1cm}

    \subfigure[\raggedright{Top-10 precision for all ridge regression models.}]{
        \includegraphics[scale=0.5]{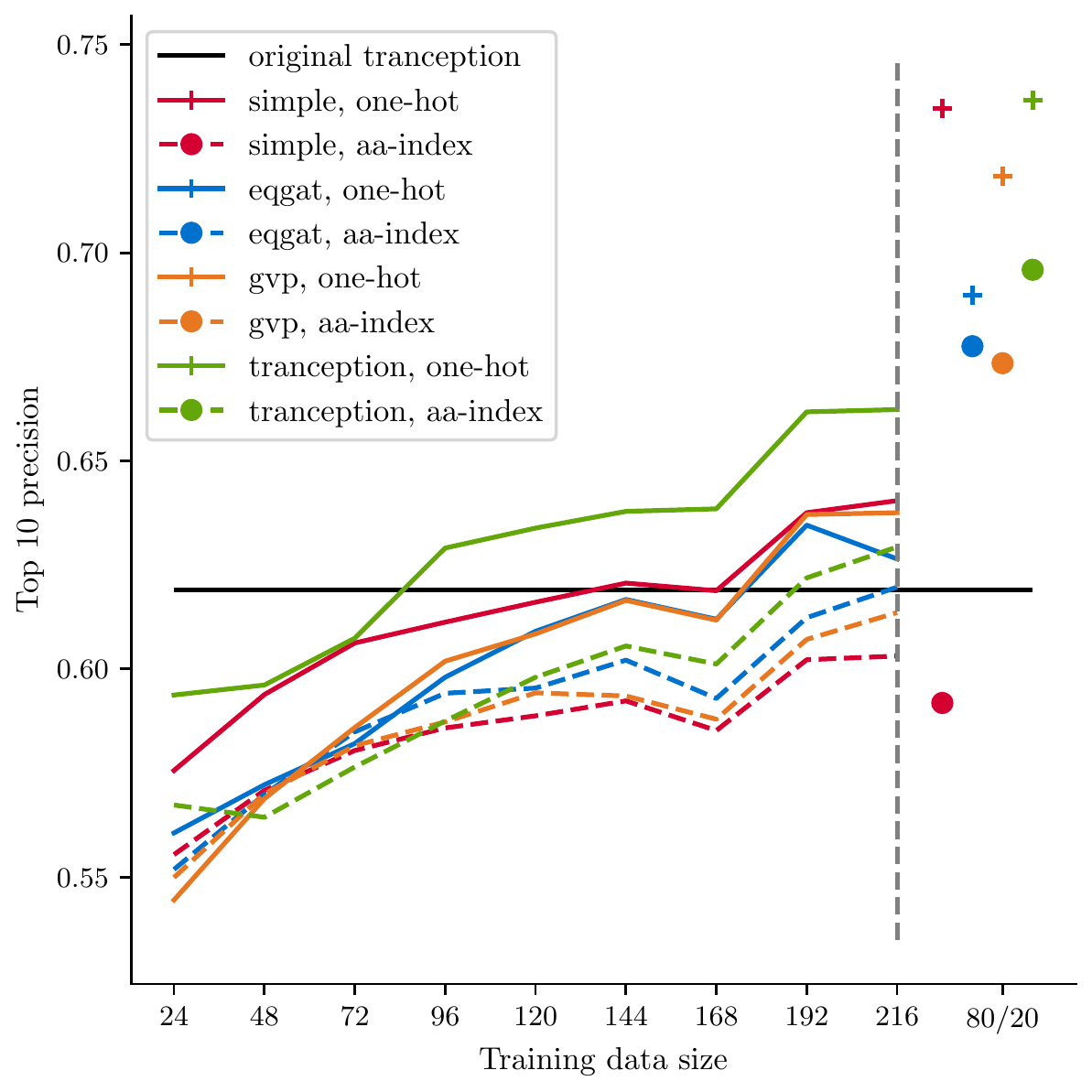}
    }
    \hspace{0.3in}
    \subfigure[\raggedright{Top-10 recall for all ridge regression models.}]{
        \includegraphics[scale=0.5]{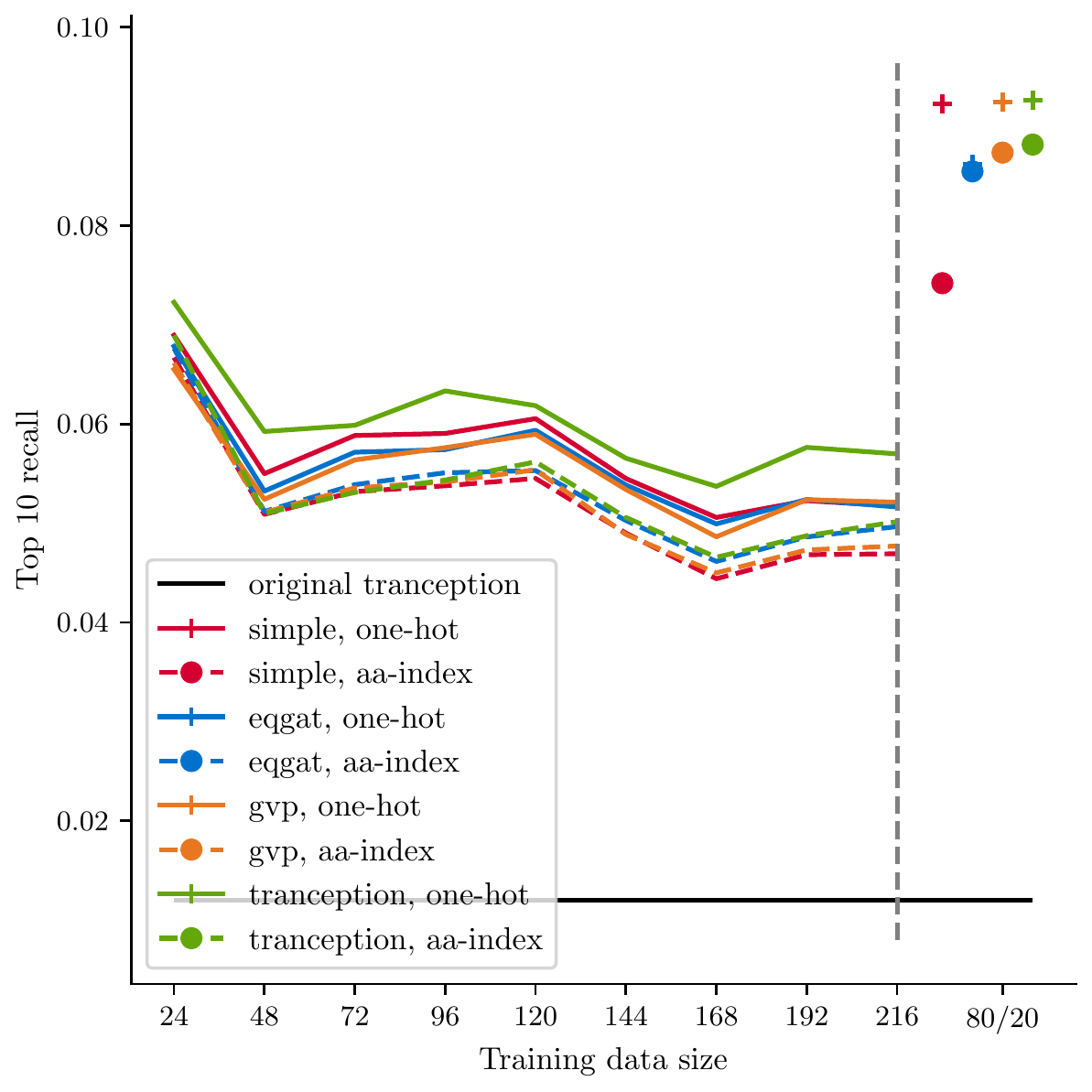}
    }
    
    \caption{}
    \label{all-performance-metrics}
\end{figure}

Figure \ref{all-performance-metrics} summarises the performance of the ridge regression models.

\newpage
\subsection{RES Task learning hyperparameters}
\label{training-details}
We train two equivariant GNN models on the RES task using the ATOM3D RES dataset \cite{atom3d}. Both models are trained using one NVIDIA A100 GPU. 

Table \ref{hyperparameters} summarises the training configuration used for both models. Table \ref{model-parameters} summarises the architectures of the EQGAT and GVP.
\begin{table}[!h]
\centering
\caption{Training configuration for both models.}
\vskip 0.15in
\begin{tabular}{@{}lc@{}}
\toprule
Hyperparameter     & Value     \\ \midrule
Learning rate      & $1e^{-4}$ \\
Patience scheduler & 10        \\
Decay rate         & 0.75      \\
Dropout            & 0.1       \\
Batch size         & 64        \\
Epochs             & 40         \\ \bottomrule
\end{tabular}
\label{hyperparameters}
\end{table}

\begin{table}[!h]
\centering
\caption{Model architectures.}
\vskip 0.15in
\begin{tabular}{@{}llc@{}}
\toprule
Model                  & Hyperparameter                 & Value     \\ \midrule
\multirow{3}{*}{GVP}   & Message-passing layers         & 5         \\
                       & Node features (scalar, vector) & (100, 16) \\
                       & Edge features (scalar, vector) & (32, 1)   \\ \midrule
\multirow{5}{*}{EQGAT} & Message-passing layers         & 5         \\
                       & Node features (scalar, vector) & (100, 16) \\
                       & RBF function                   & Bessel    \\
                       & RBFs                           & 32        \\
                       & RBF cutoff                     & 4.5 \AA   \\ \midrule

\end{tabular}
\label{model-parameters}
\end{table}

\end{document}